\documentclass[acmsmall, screen, nonacm, authorversion]{acmart}

\AtBeginDocument{%
  }

\setcopyright{acmlicensed}
\copyrightyear{2025}
\acmYear{2025}
\acmDOI{XXXXXXX.XXXXXXX}

\usepackage{latexsym} 
\usepackage{wasysym}
\usepackage{booktabs,tabularx,multirow,longtable,makecell,array,caption} 
\newcolumntype{Y}{>{\RaggedRight\arraybackslash}X}

\begin{document}

\title{From 2D to 3D Cognition: A Brief Survey of General World Models}

\author{Ningwei Xie}
\orcid{0000-0003-4001-1823}
\affiliation{
  \institution{China Mobile Research Institute}
  \city{Beijing}
  \country{China}
}

\author{Zizi Tian}
\orcid{0009-0004-2880-0963}
\affiliation{
  \institution{China Mobile Research Institute}
  \city{Beijing}
  \country{China}
}

\author{Lei Yang}
\authornote{Corresponding authors}
\affiliation{%
  \institution{China Mobile Research Institute}
  \city{Beijing}
  \country{China}}

\author{Xiao-Ping Zhang}
\authornotemark[1]
\affiliation{%
  \institution{Shenzhen Ubiquitous Data Enabling Key Lab, Shenzhen International Graduate School, Tsinghua University}
  \city{Shenzhen}
  \country{China}}

\author{Meng Guo}
\affiliation{
  \institution{China Mobile Research Institute}
  \city{Beijing}
  \country{China}
}

\author{Jie Li}
\affiliation{
  \institution{China Mobile Research Institute}
  \city{Beijing}
  \country{China}
}

\renewcommand{\shortauthors}{Xie et al.}

\begin{abstract}
  World models have garnered increasing attention in the development of artificial general intelligence (AGI), serving as computational frameworks for learning representations of the external world and forecasting future states. While early efforts focused on 2D visual perception and simulation, recent 3D-aware generative world models have demonstrated the ability to synthesize geometrically consistent, interactive 3D environments, marking a shift toward 3D spatial cognition. Despite rapid progress, the field lacks systematic analysis to categorize emerging techniques and clarify their roles in advancing 3D cognitive world models. This survey addresses this need by introducing a conceptual framework, providing a structured and forward-looking review of world models transitioning from 2D perception to 3D cognition. Within this framework, we highlight two key technological drivers, particularly advances in 3D representations and the incorporation of world knowledge, as fundamental pillars. Building on these, we dissect three core cognitive capabilities that underpin 3D world modeling: 3D physical scene generation, 3D spatial reasoning, and 3D spatial interaction. We further examine the deployment of these capabilities in real-world applications, including embodied AI, autonomous driving, digital twin, and gaming/VR. Finally, we identify challenges across data, modeling, and deployment, and outline future directions for advancing more robust and generalizable 3D world models.
\end{abstract}




\maketitle

\section{Introduction}
World Models, computational frameworks that construct representations of the external world and predict its future states, have emerged as a central research area in the pursuit of AGI. Leading AI research organizations such as Meta and OpenAI have made significant strides in this domain, as exemplified by Meta’s self-supervised JEPA architecture\cite{jepa} and OpenAI's Sora video generation model\cite{sora}. JEPA enables abstract representation learning by modeling spatiotemporal regularities in latent space, while Sora exhibits strong video generation capabilities through implicit physical dynamics modeling. More recently, 3D-aware generative world models, including World Lab's Large World Models (LWMs), Google DeepMind's Genie 2\cite{genie2}, and Odyssey's Explorer\cite{odyssey}, have extended progress into the 3D domain. By leveraging advances in neural scene representations, these models enable the generation of geometrically consistent environments with persistent objects and coherent dynamics. This collective progress signifies a paradigm shift: world models are evolving from 2D simulations toward 3D cognitive systems that can perceive, reason, and interact with complex 3D environments. Despite these advances, a comprehensive and systematic understanding of this transition from 2D perceptual to 3D cognitive world modeling remains underexplored. As research approaches an inflection point, there is an urgent need to categorize and analyze the core technical pathways, capabilities, and challenges that shape the ongoing development of 3D cognitive world models. This survey aims to fill this gap by providing a structured perspective, identifying 3D representations and world knowledge as fundamental pillars, outlining the core cognitive capabilities, and reviewing representative technical methodologies and application trajectories across key domains.

\subsection{Technological Evolution}

The concept of world models can be traced back to the theory of Mental Models in the field of cognitive science\cite{mental_model1}. Psychologist Kenneth Craik proposed that humans construct cognition and perform reasoning by abstracting the external world into simple elements and their interrelationships. J. W. Forrester, a pioneer in the field of system dynamics, applied similar principles to model complex systems, further demonstrating the utility of abstract representations in understanding and predicting system behavior\cite{mental_model2}. This theory also laid the philosophical foundation for early research on reinforcement learning and robotic control by formalizing methods for agents to learn internal models of environmental dynamics. In 2018, D. Ha and J. Schmidhuber \cite{recurrent_world_models, world_models} introduced an influential approach to world models in reinforcement learning research. Their hierarchical architecture synergistically integrates Variational Autoencoders (VAE) for spatial compression of high-dimensional perceptual inputs and Recurrent Neural Networks (RNN) for temporal dynamics modeling. Similar approaches \cite{dreamerv1, dreamerv2, transformer-based-WM, transdreamer} enabled AI agents to internally simulate environment dynamics through spatiotemporal latent representations and predict hypothetical outcomes for unseen action sequences, outperforming traditional model-based reinforcement learning algorithms in complex decision-making scenarios such as racing navigation.

In 2022, Yann LeCun proposed the Joint Embedding Predictive Architecture (JEPA), a framework inspired by biological learning mechanisms that enables machines to observe and understand the world in a manner akin to humans and animals\cite{jepa}. This architecture represents a significant advancement in the paradigm of world models through its non-generative approach to predictive learning and hierarchical decomposition of complex tasks across multiple abstraction levels and temporal scales. First, JEPA processes multi-modal sensory inputs through a joint embedding space to generate abstract environment states. The joint embedding space achieves highly efficient encoding of world dynamics, facilitating the prediction of plausible future states and robust representation learning from limited observational data. Then, JEPA models multi-scale temporal dependencies by three operational layers: high-level predictors perform long-term goal planning, mid-level components coordinate action sequences, and low-level modules execute precise motion control. This helps to resolve long-range dependency limitations in conventional models while enhancing interpretability. V-JEPA\cite{v-jepa} extends the JEPA architecture to video data, enabling hierarchical modeling of multi-scale temporal dependencies in visual representations. Another notable work in abstract reasoning within world models is Contrastive Predictive Coding\cite{cpc}, which learns compact latent representations through self-supervised contrastive learning, enabling temporal abstraction and state compression. 

Contrasting with the abstract and structured world modeling strategy, an alternative approach relies on generative AI paradigms, such as autoregressive and diffusion models, which learn world knowledge through explicit data reconstruction. These models generate multi-modal data, including text\cite{bert}, images\cite{image_generation_1,image_generation_2,image_generation_3,image_generation_4,image_generation_5}, and videos\cite{video_generation_2, video_generation_3, video_generation_4, video_generation_5, video_generation_6, video_generation_7, video_generation_8, worldgpt}. Notable examples include the Large Language Models (LLMs)\cite{gpt, gpt2, gpt3, gpt4} and Sora\cite{sora}. Autoregressive architectures based on Transformer's global attention mechanisms effectively capture long-range spatiotemporal dependencies, enabling multi-step learning and prediction of world states. Diffusion models simulate data distributions by progressively introducing noise and subsequently learning the reverse denoising process to generate high-fidelity visual content. OpenAI characterizes Sora as a world simulator, proposing its potential to advance video generation frameworks as foundational tools for constructing comprehensive world simulations. Some analyses\cite{survey_sora_agi} indicate that Sora exhibits properties of a world model, demonstrating emergent capabilities in modeling basic physical phenomena.

\subsection{From 2D Visual Representation Toward 3D Cognition}

Abstract reasoning and data-driven generation represent two primary technological approaches in world modeling, each excelling in 2D visual representation and prediction. JEPA/V-JEPA demonstrates foresight in task decomposition and logical reasoning, while models like Sora exhibit outstanding performance in 2D video fitting and generation diversity. However, both methodologies encounter limitations when addressing motion simulation, physical interaction, and causal reasoning within 3D environments. In JEPA/V-JEPA, hierarchical encoders discard low-level geometric details to enhance reasoning efficiency, which leads to degraded performance in dynamic occlusion scenarios \cite{3d_trajectory}. Sora, on the other hand, learns pixel-level correlations that lack explicit 3D structure, limiting its ability to simulate complex physical dynamics or reason about spatial causality. These observations underscore the need to develop world models that can comprehend and interact with the 3D physical world, thereby moving beyond the inherent constraints of 2D-focused approaches. 

In parallel, the rising demand for precise 3D scene construction, understanding and interaction, especially in embodied AI, autonomous driving, and digital twins, has accelerated progress in 3D generative modeling. In 2024, World Labs and Google introduced models capable of generating interactive 3D scenes from a single image\cite{genie2, odyssey, wonderworld, wonderjourney}. While these models achieve geometric plausibility, they still face challenges in delivering physically grounded semantics and responsive interactivity.  Therefore, this gap has motivated both academia and industry to pursue world models built upon explicit spatial representation and cognitive priors as a crucial next step.

\begin{figure*}[!h]
\centering
\includegraphics[width=0.75\linewidth]{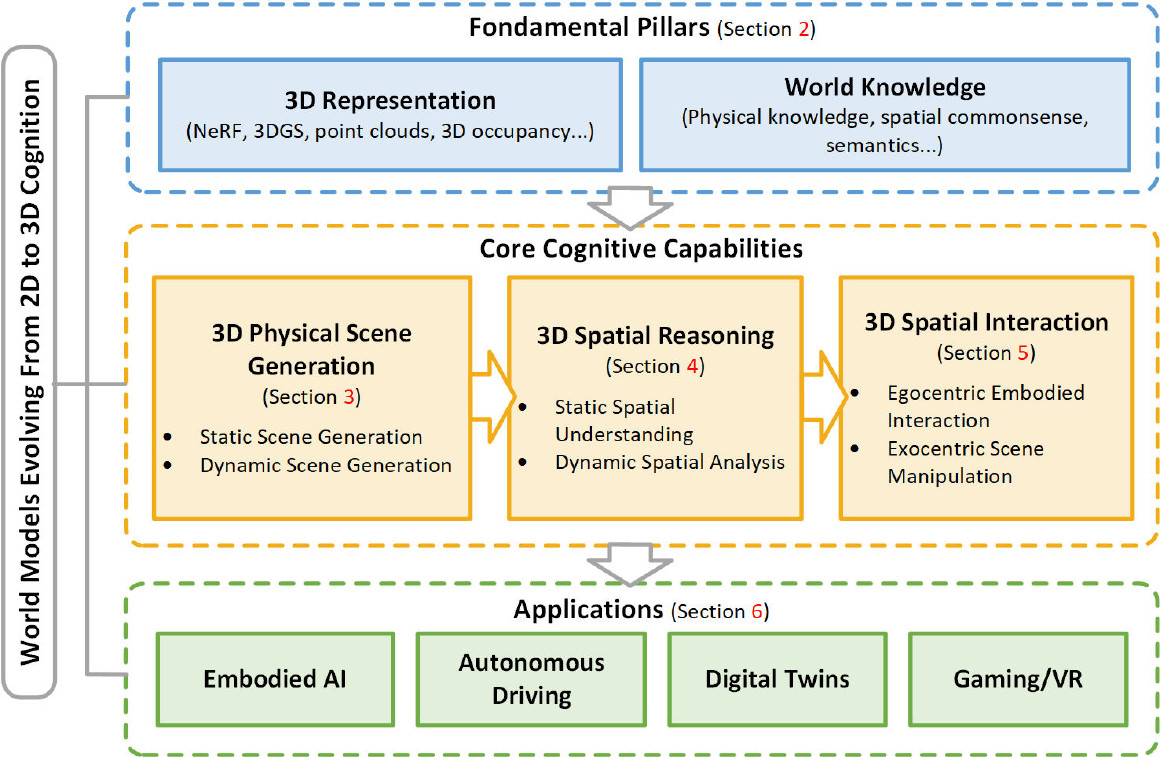}
\caption{\textbf{A conceptual framework of this survey.} Recent works advance world models toward 3D cognition by leveraging explicit 3D representations and integrating world knowledge. This facilitates the development of three fundamental cognitive capabilities within world models: 3D physical scene generation, 3D spatial reasoning, and 3D spatial interaction. As a result, these advancements significantly expand the applicability of world models across diverse domains, including embodied AI, autonomous driving, digital twins, and gaming/VR.}
\label{fig:architecture}
\end{figure*}

As illustrated in Fig. \ref{fig:architecture}, we present a conceptual framework that characterizes the development of world models around two foundational components essential for enabling spatial cognitive capabilities. The first is the adoption of \textbf{3D representations}, capturing geometry, radiance, and spatial structure in a volumetric form. The second is the incorporation of \textbf{world knowledge}, including physical laws, spatial commonsense, and semantic or structural priors, which provides contextual grounding for world modeling. Based on these two components, the framework identifies three interdependent capabilities that form the cognitive core of world models:

\begin {itemize}
\item \textbf{3D physical scene generation}, empowers models to reconstruct and synthesize volumetric environments that adhere to physical plausibility, ensuring geometric fidelity and dynamic realism. 
\item \textbf{3D spatial reasoning}, enables inference over spatial relations, object semantics or functions, and environmental dynamics, integrating geometric computation with semantic priors and commonsense knowledge to guide planning and decision making.
\item  \textbf{3D spatial interaction}, equips models with the capacity for goal-directed, physically consistent interaction, through embodied agent behaviors and user-driven editing, transforming them from passive observers to active participants in 3D environments.
\end {itemize}

This capability triad mirrors the canonical perceive–think–act loop commonly used to describe intelligent systems\cite{russell2021aimodern}, and also serves as an organizing principle for surveying and evaluating recent progress across four key application domains, including embodied AI, digital twins, autonomous driving, and gaming/VR.

\subsection{Comparison with Prior Surveys}

\setlength{\tabcolsep}{3pt}
\renewcommand{\arraystretch}{1.1}

\begin{table*}[!h]
  \centering
  \caption{Comparison of our survey with previous surveys in terms of main topic, and emphasis.}
  \label{tab:compare}
  \scriptsize
  \begin{tabular*}{\textwidth}{@{\extracolsep{\fill}}%
      m{0.12\textwidth}
      m{0.05\textwidth}
      *{5}{m{0.05\textwidth}}
      m{0.45\textwidth}
    @{}}
    \toprule
    \multirow{3}{*}{\textbf{Survey}} & \multirow{3}{*}{\textbf{Year}} & \multicolumn{5}{c}{\textbf{Main Topic}} & \multirow{3}{*}{\textbf{Emphasis and Scope}} \\ 
    \cmidrule(l){3-7}
    & & \textbf{3D Rep.} & \textbf{World Know.} & \textbf{Gen.} & \textbf{Reas.} & \textbf{Inter.} & \\
    \midrule
    Mai et al.\cite{efficient_multimodal} 
      & 2024 
      & \Circle & \Circle & \RIGHTcircle & \RIGHTcircle & \Circle 
      & Emphasizes cross-modal integration and language grounding; provides limited discussion on 3D generation and physical priors. \\ 
    \midrule
    Lin et al.\cite{physics_cognition} 
      & 2025 
      & \RIGHTcircle & \CIRCLE & \CIRCLE & \RIGHTcircle & \Circle
      & Focuses on video generation under physical constraints; less emphasis on 3D spatial reasoning or interaction. \\ 
    \midrule
    Hu et al.\cite{simulate_realworld}  
      & 2025 
      & \CIRCLE & \RIGHTcircle & \CIRCLE & \Circle & \Circle 
      & Highlights progression of data dimensionality (2D, video, 3D, 4D) in multimodal generative models; does not explicitly explore active spatial interaction. \\ 
    \midrule
    Ding et al.\cite{tsinghua_world_models} 
      & 2024 
      & \RIGHTcircle & \CIRCLE & \Circle & \CIRCLE & \Circle  
      & Reviews general world modeling paradigms (state representation vs. future prediction); uplaces less focus on explicit 3D representation and active interaction. \\ 
    \midrule
    \textbf{Ours} 
      & \textbf{2025} 
      & \CIRCLE & \CIRCLE & \CIRCLE & \CIRCLE & \CIRCLE 
      & Provides a structured review of general world models evolving towards 3D cognition, unifying 3D representations, world priors, and spatial cognitive capabilities across generation, reasoning, and interaction.\\ 
    \bottomrule
  \end{tabular*}
  {\raggedright\footnotesize * “\CIRCLE”, “\RIGHTcircle” and “\Circle” indicate comprehensive, partial, and no coverage, respectively. \par}
  {\raggedright\footnotesize * Columns correspond to: \textbf{3D Rep.} = 3D representation, \textbf{World Know.} = world knowledge, \textbf{Gen.} = 3D physical scene generation, \textbf{Reas.} = 3D spatial reasoning, \textbf{Inter.} = 3D spatial interaction. \par}
\end{table*}

To highlight the unique perspective of our survey, Table~\ref{tab:compare} compares our survey with prior works in terms of foundational focus, capability coverage, and thematic emphasis. While these works have advanced understanding in specific subdomains, such as multimodal learning\cite{efficient_multimodal}, physical video synthesis\cite{physics_cognition}, and dimensional scaling in generative models\cite{simulate_realworld}, they typically focus on either rendering realism or abstract structural modeling. Ding et al. (2024)\cite{tsinghua_world_models} categorize world models by state representation versus future prediction, while place less focus on explicit volumetric representations or active interactivity. In contrast, our survey provides a unified and comprehensive perspective on world models evolving towards 3D cognition. We systematically analyze recent advances through the lens of two foundational pillars, 3D representations and world knowledge, and describe how these foundations support three core spatial capabilities: 3D physical scene generation, 3D spatial reasoning, and 3D spatial interaction. 

\subsection{Structure of This Paper}

To systematically explore the components and capabilities outlined above, the remainder of this paper is organized as follows. Section~\ref{sec:fundamental} presents the two fundamental pillars of 3D world modeling: 3D representations and world knowledge. Section~\ref{sec:generation}, Section~\ref{sec:reasoning}, and Section~\ref{sec:interaction} respectively further discuss the three core cognitive capabilities: 3D physical generation, 3D spatial reasoning, and 3D spatial interaction. Section~\ref{sec:application} illustrates how these capabilities are applied across key domains, including embodied AI, autonomous driving, digital twins, and gaming/VR. Section~\ref{sec:challenge} concludes with an overview of open challenges and promising directions for advancing toward general 3D world cognition.

\section{Fundamental Pillars}
\label{sec:fundamental}

World models with 3D cognition  are fundamentally grounded in 3D representations and enriched by world knowledge. This section overviews recent advances in 3D representations and introduces two key types of world knowledge: physical knowledge from physics simulation and learned knowledge via large pretrained models, forming the basis for the 3D cognitive capabilities discussed later.

\subsection{3D Representations}

\begin{figure*}[!t]
\centering
\includegraphics[width=0.9\linewidth]{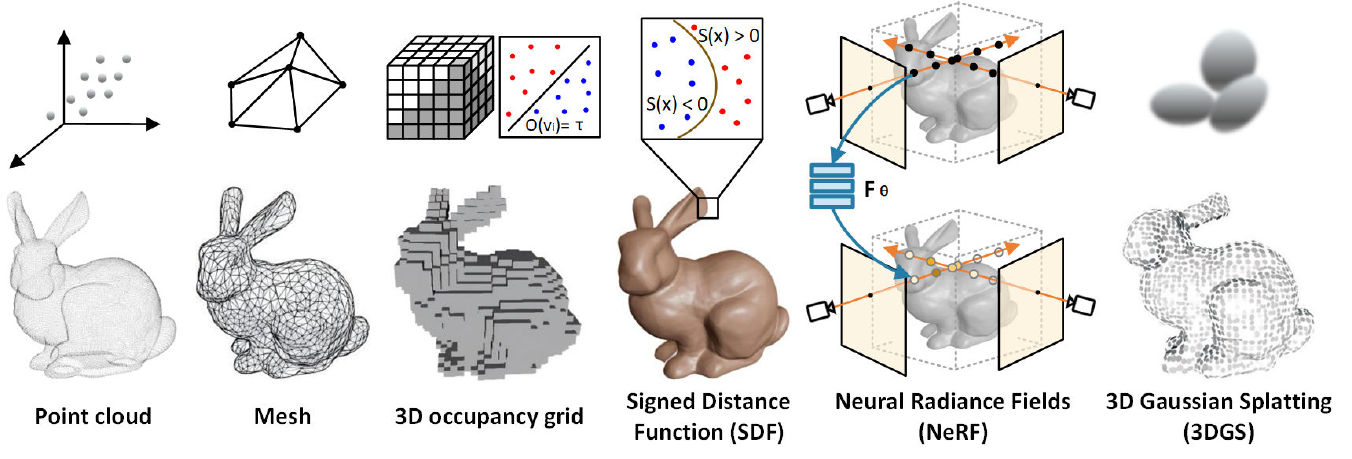}
\caption{\textbf{3D representations} that are covered in this
survey, including point cloud, mesh, 3D occupancy grid, SDF, NeRF, 3DGS. }
\label{fig:representation}
\end{figure*}

We introduce 3D representations including point cloud, mesh, 3D occupancy grid, Signed Distance Function (SDF)\cite{deepsdf}, Neural Radiance Fields (NeRF)\cite{nerf} and 3D Gaussian Splatting (3DGS)\cite{gaussian}. We provide a schematic illustration in Fig.~\ref{fig:representation}. 

\subsubsection{Point cloud} Point cloud is a widely adopted representation of 3D data, consisting of discrete sets of spatial points. Point clouds directly reflect the raw output from range sensors such as LiDARs and depth cameras, and offer an lightweight and flexible format for representing large-scale 3D scenes. PointNet\cite{pointnet, pointnet++} pioneered learning directly on point sets, while more recent models like Point Transformer\cite{pointtransformer} introduce attention-based mechanisms to enhance point cloud feature extraction. These models have become backbones for a wide range of 3D understanding tasks.

\subsubsection{Mesh} Mesh comprises vertices, edges, and faces that define a 3D shape's geometry and topology. Owing to its surface-aware structure and high geometric fidelity, mesh is widely adopted as output representations in 3D reconstruction tasks\cite{meshreconstruct1, meshreconstruct2}. Moreover, it is well-suited for parameterized, template-based modeling of dynamic objects. For example, a 3D morphable model of the human body, SMPL\cite{smpl}, deforms a canonical mesh based on learned shape and pose parameters to capture articulated motion. 

\subsubsection{3D occupancy grid} 3D occupancy grid divides space into a regular voxel grid, where each voxel stores a binary or probabilistic value indicating whether its spatial location is occupied. Formally, an occupancy field is defined as $\mathcal{O} : \mathbb{R}^3 \rightarrow [0,1]$. $\mathcal{O}(v_i)$ denoting the occupancy probability of a voxel centered at $v_i$. Here, $\mathcal{O}(v_i) = 1$ indicates full occupancy, while $\mathcal{O}(v_i) = 0$ represents free space. This structured representation supports spatial reasoning and incremental scene construction from sensor data. Beyond binary encoding, semantic labels can be integrated into voxels to jointly capture spatial structure and scene semantics~\cite{occ3d}. Notably, Occupancy Networks~\cite{occupancynetworks} generalize this representation by learning a continuous occupancy field with neural networks, enabling multi-resolution mesh extraction.

\subsubsection{Signed Distance Function (SDF)} SDF is an implicit 3D representation that defines a continuous scalar field over space, assigning each point a signed distance to the nearest surface. Formally, $\mathcal{S} : \mathbb{R}^3 \rightarrow \mathbb{R}$, where $\mathcal{S}(x) > 0$ denotes points outside the surface, $\mathcal{S}(x) < 0$ inside, and $\mathcal{S}(x) = 0$ on the surface. A representative method, DeepSDF~\cite{deepsdf}, learns such implicit surfaces using deep neural networks, predicting signed distances from input coordinates and latent shape codes to enable high-fidelity reconstruction and shape interpolation.

\subsubsection{Neural Radiance Fields (NeRF)} NeRF models 3D scenes as continuous volumetric fields by learning a function $\mathcal{F}$ that maps a 3D position $\mathbf{x} \in \mathbb{R}^3$ and viewing direction $\mathbf{d} \in \mathbb{R}^2$ to an RGB color $\mathbf{c} \in \mathbb{R}^3$ and volume density $\sigma$, as defined in Eq.\ref{eq:nerf}.
\begin{equation}
\mathcal{F}_\theta : (\mathbf{x}, \mathbf{d}) \rightarrow (\mathbf{c}, \sigma).
\label{eq:nerf}
\end{equation}
This mapping is typically parameterized by a multilayer perceptron (MLP) with parameters $\theta$. Novel views are synthesized through volumetric rendering by integrating color along camera rays. NeRF enables photorealistic view synthesis from sparse inputs, and has been widely adopted for high-fidelity static scene reconstruction.

\subsubsection{3D Gaussian Splatting (3DGS)} 3DGS explicitly encodes a radiance field using a set of Gaussian kernels $\mathcal{P}$, where each kernel is defined by its center position ${\bf{x}}_p$ and covariance matrices ${\bf{A}}_p$, and is associated with an opacity ${o_p}$ and a view-dependent color function ${\bf{c}}_p$, $p \in \mathcal{P}$\cite{gaussian}. Given a rendering viewpoint, the Gaussian ellipsoids are projected onto the imaging plane, forming 2D Gaussian splats. Formally, the $k$-th pixel color is computed as Eq.~\ref{eq:gaussian}.
\begin{equation}
{c_k} = \sum\limits_{i = 1}^N {{G_i}(k){o_i}{{\bf{c}}_i}} ({{\bf{r}}_k})\prod\limits_{j = 1}^{i - 1} {(1 - {G_j}(k){o_j}).}
\label{eq:gaussian}
\end{equation}
${G_i}(k)$ represents the 2D Gaussian weight of the $i$-th kernel at pixel $k$, ${{\bf{r}}_k}$ denotes of the viewing direction of the camera. 3DGS supports direct optimization of point attributes and enables real-time rendering. 

Recent extensions of NeRF and 3DGS enable dynamic scene modeling\cite{dynamic-synthesis, d-nerf, nerfies, fast-nerf, physgaussian, deformgs}, or integrate with diffusion models for photorealistic 3D content creation\cite{dreamfusion, magic3d, prolificdreamer, luciddreamer, dreamgaussian, latentnerf, zero-1-to-3}, highlighting their representational flexibility.

\subsection{Physical Knowledge from Physics Simulation} 

\begin{figure*}[!t]
\centering
\includegraphics[width=0.75\linewidth]{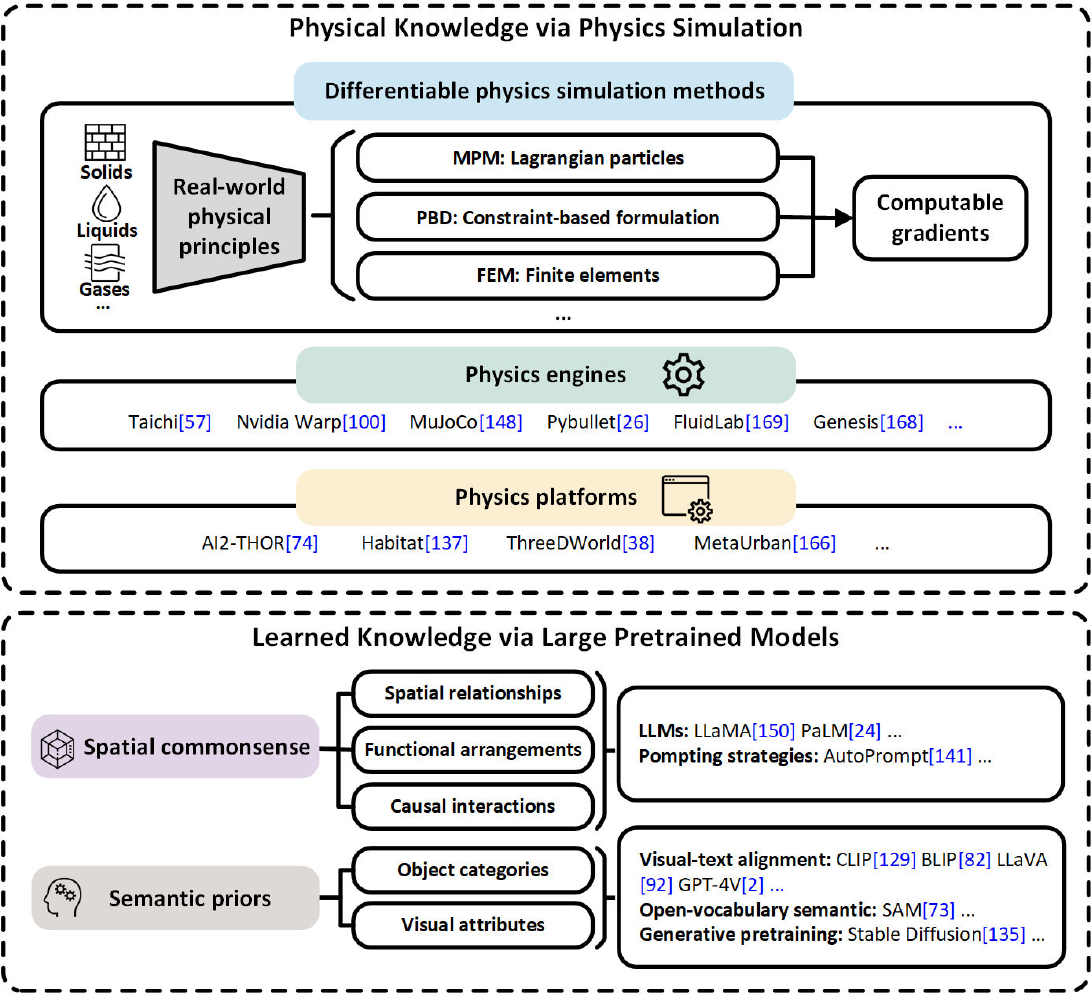}
\caption{\textbf{Illustration of world knowledge sources}. Top: Real-world physical laws are encoded via differentiable simulation techniques. Modern engines and platforms provide various simulation functionalities, supporting physical world modeling. Bottom: Learned knowledge from large pretrained models captures spatial commonsense and semantic priors, including relational, functional, and categorical cues.}
\label{fig:knowledge}
\end{figure*}

Physics simulation is fundamental to physical world modeling, as it encodes physical laws into computable signals that can be incorporated into world model optimization\cite{end2end, accelerate_policy, diffsim}. We introduce classic simulation methods and supporting platforms that serve as the computational backbone for integrating physical priors into world models, as illustrated by the upper part of Fig.~\ref{fig:knowledge}.

\subsubsection{Physics simulation methods} Three major methods have been widely explored in this context: 
\begin{itemize}
\item \textbf{Material Point Method (MPM)}\cite{mpm} discretizes materials into a set of Lagrangian particles that carry physical quantities such as mass, velocity, and stress. These particles transfer information to a background Eulerian grid, where Newton’s laws of motion are solved to compute forces and update momentum. The updated values are then mapped back to the particles, enabling stable simulation of large deformations and collisions.
\item \textbf{Position-Based Dynamics (PBD)}\cite{pbd, xpbd} iteratively updates particle positions under physical constraint conditions, rather than computing forces explicitly. This constraint-based formulation enables stable and efficient simulation of soft-body dynamics and fluid-like behaviors.
\item \textbf{Finite Element Method (FEM)}\cite{fem} models physical systems by discretizing a continuous domain into a mesh of finite elements, within which partial differential equations are solved to simulate physical responses. It is particularly effective in modeling elastic and plastic deformations under complex boundary conditions.
\end{itemize}

\subsubsection{Physics engines and platforms} Modern differentiable physics engines are built on GPU-accelerated, gradient-enabled computational frameworks, such as Taichi\cite{taichi} and Nvidia Warp\cite{warp}, and support simulation of complex material behaviors\cite{difftaichi, nimblephysics, diffskill, dojo, softzoo, plasticinelab, phiflow, jaxfluids, mujoco, pybullet}. Notably, FluidLab\cite{fluidlab} builds a unified simulation environment supporting solids, liquids, gases, and their interactions, enabling multi-material differentiation with high efficiency. Genesis\cite{genesis} further extends differentiable physics into a generative platform that translates natural language prompts into physically plausible simulation sequences, demonstrating potential for general-purpose, user-controllable physics simulation.

Large-scale 3D simulation platforms provide physics-rich, multi-modal, and interactive environments that serve as canonical sources or testbeds for training and evaluating world models. By providing high‑fidelity ground‑truth state transitions and observations, these platforms underpin reliable sim‑to‑real transfer. AI2-THOR~\cite{ai2thor} simulates realistic indoor scenes with rich object manipulation, scene rearrangement, and step-wise metadata. Habitat~\cite{habitat} enables efficient rendering of building-scale environments at thousands of frames per second with realistic lighting and occlusion. ThreeDWorld\cite{tdw} simulates multi-modal sensory streams, including RGB, depth, and audio, alongside high-fidelity physics for soft bodies, liquids, and deformable objects. MetaUrban\cite{metaurban} extends embodied simulation to outdoor, city-scale environments via a hierarchical procedural pipeline generating infinite street-block layouts, functional zones, and dynamic entities.

Together, these simulation techniques and platforms provide foundational physical knowledge for world models, improving the realism of geometry and material representations in static scenes, and ensuring physically consistent motion and deformation in dynamic environments.

\subsection{Learned Knowledge via Large Pretrained Models}

Recent advances in large pretrained models, especially LLMs and Vision-Language Models (VLMs), have enabled the extraction of high-level world knowledge from massive textual and visual corpora. We focus on two primary categories of learned knowledge that world models benefit from: spatial commonsense and semantic priors.

\subsubsection{Spatial commonsense} Spatial commonsense refers to intuitive knowledge about typical spatial relationships, functional arrangements and causal interactions of objects in the physical world, providing contextual concepts such as "a cup is typically found on a table," "a bed is placed along a wall". Recent studies show that pretrained LLMs, such as LLaMA\cite{llama} and PaLM\cite{palm} implicitly encode such spatial and causal knowledge. These models perform competitively on spatial commonsense benchmarks like PIQA\cite{piqa}, which test models’ abilities to infer plausible physical interactions. Moreover, the non-multimodal version of GPT-4V\cite{gpt4} exhibit emergent capabilities in interpreting the 3D environments through codes\cite{gpt4-experiment}. In parallel, some methods propose to elicit knowledge from pretrained LLMs by introducing prompting engineering strategies. In parallel, to more effectively elicit knowledge encoded by LLMs, various prompting strategies have been proposed. For instance, AutoPrompt\cite{autoprompt} automates the generation of task-specific trigger tokens that activate latent knowledge in LLMs. Chain-of-thought prompting\cite{cot} facilitates the transfer of LLMs to tasks requiring hierarchical spatial and causal reasoning by decomposing complex problems into structured intermediate steps.

\subsubsection{Semantic priors} Semantic priors refer to knowledge about object categories and their associated visual attributes. These priors can be typically learned by powerful VLMs pretrained on large-scale visual-textual datasets\cite{vlm_survey}. Notably, the text-image alignment methods, represented by CLIP\cite{clip}, link textual concepts with visual evidence by aligning textual and visual embeddings in a shared latent space via contrastive learning, providing strong semantic discriminability for object categories and attributes. Recent models such as BLIP\cite{blip, blip-2}, GPT-4V\cite{gpt4}, LLaVA\cite{llava} enhance abilities in encoding compositional semantics to support relational reasoning and context-aware visual understanding. Open-vocabulary grounding and segmentation models\cite{sam, dino, grounding-dino, groundedsam, lisa} have demonstrated impressive generalization to novel object categories and spatial concepts without requiring retraining. These models provide category-agnostic proposals and fine-grained pixel-level masks, enabling world models to incorporate high-level semantics at the object, region, and scene levels, which supports more informed scene understanding and interaction planning. Furthermore, 2D diffusion models such as Stable Diffusion\cite{stablediffusion} possess strong language-vision alignment through generative pretraining on text-image pairs. These models offer semantic-rich visual priors and have been leveraged to supervising 3D representations for controllable and semantically aligned 3D scene generation. 

As elaborated in the following sections, world models achieves generalized, interpretable spatial understanding and inference by querying or conditioning 3D representations on contextual and semantic information derived from these foundation models.

\section{3D Physical Scene Generation}
\label{sec:generation}

Generating physically plausible 3D environments forms the perceptual foundation of world models with genuine 3D cognition. Unlike traditional 3D scene synthesis techniques that prioritize visual fidelity, this capability emphasizes geometric consistency, material plausibility, and physical coherence, ensuring that generated scenes are not only photorealistic but also adhere to the structural and dynamical principles of the real world. To this end, recent research trends increasingly advocate for the integration of world priors into 3D scene generation frameworks.

In this section, we classify 3D physical scene generation methods into two major categories: static scene generation and dynamic scene generation.

\subsection{Static Scene Generation}

\renewcommand{\arraystretch}{1.1}  
\setlength{\tabcolsep}{3pt}         
\begin{table*}[!t]
  \centering
  \caption{Overview of representative works for static physical scene generation.}
  \label{tab:static_generation}
  \scriptsize
  \begin{tabular*}{\textwidth}{@{\extracolsep{\fill}}%
      m{0.07\textwidth}
      m{0.18\textwidth}
      m{0.13\textwidth}
      m{0.27\textwidth}
      m{0.23\textwidth}
    @{}}
    \toprule
    \textbf{Category} & \textbf{Method} & \textbf{3D Rep.} & \textbf{Priors} & \textbf{Applicability} \\
    \midrule
    \multirow{5}{=}{Layout-guided}
      & LI3D~\cite{li3d} & NeRF & \multirow{5}{=}{LLM-inferred layouts (object categories, location, size, orientation, appearance, etc.)} & \multirow{5}{=}{Scenes conforming to spatial commonsense and semantic consistency} \\
    \cmidrule{2-3}
      & SceneTeller~\cite{sceneteller} & 3DGS &  &  \\
    \cmidrule{2-3}
      & GALA3D~\cite{gala3d}    & 3DGS &  &  \\
    \cmidrule{2-3}
      & SceneWiz3D~\cite{scenewiz3d} & NeRF &  &  \\
    \midrule
    \multirow{7}{=}{Physics-informed}
      & LayoutDreamer~\cite{layoutdreamer} & 3DGS & Physical constraints (gravity, center-of-mass stability, inter-object penetration) & \multirow{7}{=}{Physically plausible and stable sceness} \\
    \cmidrule{2-4}
      & PhyScene~\cite{physcene} & Latent 3D features & Physical constraints (reachability, collision-avoidance) &  \\
    \cmidrule{2-4}
      & Phy3DGen~\cite{phy3dgen}    & Mesh \& SDF & Physical priors (stress, strain, displacement) &  \\
    \cmidrule{2-4}
      & PhysComp3D~\cite{physcomp3d} & Mesh & Material priors (mechanical responses under external forces) &  \\
    \bottomrule
  \end{tabular*}
  \vspace{1ex}
  {\raggedright\footnotesize * Abbreviations: 3D Rep.\ = 3D Representation.\par}
\end{table*}

This category focuses on methods that integrate structural and physical priors to generate 3D static scenes. Table~\ref{tab:static_generation} summarizes the representative methods, along with their 3D representations, incorporated priors or constraints and targeted applicability.

\subsubsection{Layout-guided generation} Early approaches guide object placement (including position, rotation, and scale) and scene composition using manually designed layout priors, to ensure spatial consistency and controllability. Set-the-Scene\cite{set-the-scene} allows users to place layout anchors that serve as global guidance for training locally controlled NeRFs, enabling controllable and globally consistent scene generation. Similarly, Comp3D\cite{comp3d} introduces a locally conditioned diffusion framework, where user-defined bounding boxes and corresponding text prompts jointly guide the generation of 3D scene components. Then, these components are then assembled into a coherent voxel-based NeRF representation, achieving object-level control and high spatial fidelity.

Moving beyond hand-crafted priors, recent advances enhance scene layout learning by implicitly leveraging commonsense spatial knowledge encoded in LLMs. LI3D\cite{li3d} treats LLMs as interpreters of layout instructions and integrates generative feedback mechanisms to iteratively refine the 3D scene generated by CompoNeRF\cite{componerf}. By exploiting LLM's pre‑trained understanding of spatial relationships, the framework infers plausible coarse layouts, ultimately producing semantically consistent and visually realistic 3D scenes. SceneTeller\cite{sceneteller} further leverages LLMs with in-context learning to parse textual descriptions into a set of 3D bounding boxes, each specifying an object’s type, position, size, and orientation. The derived layout is subsequently fitted with a 3DGS representation. GALA3D\cite{gala3d} enhances LLM-generated initial layouts by introducing an adaptive geometric control mechanism. It aligns 3D Gaussians with text-to-image diffusion priors through compositional optimization, ensuring high geometric and textural consistency in multi-object scenes. SceneWiz3D\cite{scenewiz3d} utilizes LLMs to extract objects of interest (OOI) from input texts, initializes their 3D representations via text-to-3D retrieval. It also employs particle swarm optimization (PSO)\cite{pso} based on CLIP\cite{clip} similarity to automatically configure object placement, thereby satisfying both semantic and spatial consistencies.

\subsubsection{Physics-informed generation} In addition to structural and commonsense priors, some approaches incorporate explicit physical knowledge to ensure the generated layouts are physically valid. LayoutDreamer\cite{layoutdreamer} introduces a physics-guided hierarchical layout optimization framework. It formulates physical knowledge as energy terms, including gravity, center-of-mass stability, anchoring, and inter-object penetration, in a two-stage optimization pipeline over a 3D Gaussian scene representation. By minimizing these energy functions, LayoutDreamer ensures that the resulting layouts are both physically stable and semantically coherent. PhyScene\cite{physcene} proposes a diffusion-based generation framework augmented with differentiable physical constraint functions. It integrates layout regularization, collision avoidance, and reachability optimization into the denoising process, enabling the creation of physically interactable static scenes. 

Moreover, several approaches focus on improving the physical realism of individual object for compositional scene generation. Phy3DGen\cite{phy3dgen} introduces a differentiable physical layer inspired by the FEM, which models solid mechanics using three key physical quantities, i.e., stress, strain, and displacement. This is incorporated into a text-to-3D diffusion model to jointly optimize the visual fidelity and physical accuracy. Specifically, Phy3DGen can generate 3D shapes with more uniform stress distribution, which conform to the principles of solid mechanics. PhysComp3D\cite{physcomp3d} proposes physical compatibility optimization, which refines the rest-shape geometry of 3D objects by modeling mechanical material properties and external forces. These precise-physics‑driven approaches effectively produce static scenes capable of exhibiting accurate mechanical responses under load, enabling seamless downstream dynamic simulation.

\subsection{Dynamic Scene Generation}

\setlength{\tabcolsep}{3pt}           
\renewcommand{\arraystretch}{1.1}     

\begin{table*}[!t]
  \centering
  \caption{Overview of representative works for dynamic physical scene generation.}
  \label{tab:dynamic_generation}
  \scriptsize
  \begin{tabular*}{\textwidth}{@{\extracolsep{\fill}}%
      m{0.13\textwidth}
      m{0.17\textwidth}
      m{0.06\textwidth}
      m{0.12\textwidth}
      m{0.22\textwidth}
      m{0.18\textwidth}
    @{}}
    \toprule
    \textbf{Category} & \textbf{Method} & \textbf{3D Rep.} 
      & \textbf{Physics Simulator} & \textbf{Priors} & \textbf{Applicability} \\
    \midrule

    \multirow{3}{=}{Physics-regularized}
      & GausSim~\cite{gaussim}       & 3DGS  & —   & Mass \& momentum conservation regularization & Elastic objects \\
    \cmidrule{2-6}
      & DeformGS~\cite{deformgs}     & 3DGS  & —   & Momentum \& local isometry regularization      & Large‑deformable objects \\
    \midrule

    \multirow{8}{=}{Physics simulation‑based}
      & ParticleNeRF~\cite{particlenerf}      & NeRF  & PBD & Collision‑avoidance constraint                 & Rigid, articulated, deformable objects \\
    \cmidrule{2-6}
      & PIE-NeRF~\cite{pie-nerf}              & NeRF  & Q‑GMLS & Elastodynamics                           & Large‑deformable objects \\
    \cmidrule{2-6}
      & PhysGaussian~\cite{physgaussian}     & 3DGS  & MPM & Continuum mechanics                        & Diverse materials \\
    \cmidrule{2-6}
      & Phy124~\cite{phy124}                 & 3DGS  & MPM & Continuum mechanics                        & Elastic objects \\
    \cmidrule{2-6}
      & Gaussian Splashing~\cite{gaussiansplashing} & 3DGS  & PBD & Surface tension, fluid reflection/refraction effects & Solid‑fluid interactions \\
    \midrule

    \multirow{11}{=}{Physics simulation‑based + learnable physical properties}
      & PAC‑NeRF~\cite{pac-nerf}         & NeRF  & MPM & Continuum mechanics                          & Diverse materials \\
    \cmidrule{2-6}
      & DANO~\cite{dano}                & NeRF  & Dojo~\cite{dojo} & Rigid‑body contact model        & Rigid objects \\
    \cmidrule{2-6}
      & Spring‑Gaus~\cite{springgaus}   & 3DGS  & Custom spring-mass simulator & Elastic dynamics            & Elastic objects \\
    \cmidrule{2-6}
      & NeuMa~\cite{neuma}              & 3DGS  & MPM & Expert‑designed constitutive models           & Diverse materials \\
    \cmidrule{2-6}
      & OmniPhysGS~\cite{omniphysgs}    & 3DGS  & MPM & Expert‑designed constitutive models           & Diverse materials \\
    \cmidrule{2-6}
      & PhysSplat~\cite{physplat}      & 3DGS  & MPM & Physical commonsense via pretrained LLMs and VLMs & Diverse materials \\
    \bottomrule
  \end{tabular*}
  {\raggedright\footnotesize * Abbreviations: 3D Rep. = 3D Representation.\par}
\end{table*}

This category focuses on methods that generate dynamic scenes under physical constraints, capturing object deformation, motion, and contact dynamics. An overview of representative methods and their physical modeling characteristics is provided in Table~\ref{tab:dynamic_generation}. 

\subsubsection{Physics-regularized generation} Some methods introduce explicit physics regularization to optimize physically plausible dynamics. GausSim\cite{gaussim} treats each 3D Gaussian kernel as a continuous piece of matter organized into explicit Center of Mass Systems (CMS). Leveraging explicit continuum‑mechanics constraints, it effectively captures realistic elastic deformations of real‑world objects. DeformGS\cite{deformgs} employs physics regularization terms, such as conservation of momentum and local isometry, to guide the deformation of 3D Gaussians. Concurrently, it utilizes a learnable deformation function, implemented via a multilayer perceptron (MLP) with neural-voxel encoding, to map canonical Gaussian properties to their deformed states in world space. This combination allows DeformGS to effectively model scene flow in highly deformable scenes, balancing physical plausibility with the flexibility of data-driven learning.

\subsubsection{Physics simulation-based generation} A growling line of methods integrate physics simulatiors with NeRF and 3D Gaussian representations to model dynamic phenomena. In the NeRF setting, ParticleNeRF\cite{particlenerf} introduces a particle-based encoding scheme for online NeRF optimization, where each particle carries local appearance and geometric features. It incorporates a PBD module into the NeRF formulation to update the motion of the particles under collision-avoidance constraint. PIE-NeRF\cite{pie-nerf} explicitly integrates elastodynamic simulation with NeRF, enabling photorealistic rendering of soft object deformations under external force. It introduces a mesh-free simulation scheme based on Quadratic Generalized Moving Least Squares (Q-GMLS), supporting real-time computation of nonlinear elastic responses governed by the material models. By adapting the simulation kernels to the NeRF density distribution, the framework enables large-deformation modeling and efficient force propagation within the volumetric NeRF representation.

Due to their particle-based nature, 3D Gaussians can be endowed with physical properties and seamlessly integrated with physics simulators. PhysGaussian\cite{physgaussian} and Phy124\cite{phy124} extends each Gaussian with physical attributes based on a custom MPM, allowing the Gaussian kernels and their associated spherical harmonics to evolve according to the laws of continuum mechanics. These Gaussians act simultaneously as simulation particles and as renderable primitives. This design ensures consistency between physical dynamics and rendered appearance. Gaussian Splashing\cite{gaussiansplashing} generate fluid-solid dynamics within 3DGS scenes through PBD simulation. It separates solid simulation and rendering by simulating sampled particles and interpolating their deformations onto trained Gaussian kernels for rendering. For fluid simulation and rendering, it uses a unified set of Gaussian kernels and optimizes them to produce plausible diffuse, specular shadings and surface roughness.

Beyond direct simulation, some approaches aim to learn or adapt physical properties for specific materials. PAC-NeRF\cite{pac-nerf} employs a hybrid Eulerian–Lagrangian approach, where static geometry is encoded via a voxel-based Eulerian NeRF, while dynamics are modeled using MPM-driven Lagrangian particles. Notably, it learns unknown geometry and physical parameters from multi-view videos, enabling accurate dynamic modeling across diverse materials, including elastic solids, plasticine, and fluids. DANO\cite{dano} estimates inertial and friction properties and integrates the differentiable physics simulator Dojo\cite{dojo} with NeRFs to model rigid-body contact dynamics. Spring-Gaus\cite{springgaus} develops a learnable, differentiable spring-mass simulator tailored for 3D Gaussians, jointly learning elasticity parameters and dynamic motion directly from multi-view video. Each Gaussian serves as a mass node connected by virtual springs, enabling physically consistent elastic dynamics via position updates governed by Hooke’s law. NeuMa\cite{neuma} learns to correct expert-designed physical models by fitting a neural material adaptor to ground-truth visual observations. It predicts latent material embeddings that capture actual intrinsic dynamics, to guide the generating of physically plausible trajectory and deformation of objects. OmniPhysGS\cite{omniphysgs} employs several learnable constitutive models to predict material responses for each Gaussian under specific mechanical conditions, which are then used to drive physical simulation within a differentiable MPM module. This design allows OmniPhysGS to model a wide range of material behaviors, including elasticity, plasticity, and viscoelasticity, within a unified generative framework. 

Different from these constitutive learning approaches, another line of work leverages LLMs to infer physical properties from visual observations. For example, PhysSplat\cite{physplat} designs a multi-modal LLMs (BLIP, CLIP, GPT-4V) to predict material categories and physical properties of dynamic objects based on input video. A Material Property Distribution Prediction (MPDP) module models plausible distributions over physical parameters, which are then used in a differentiable MPM simulator to generate physically consistent and generalizable dynamic scene synthesis.

\begin{figure*}[!t]
\centering
\includegraphics[width=0.8\linewidth]{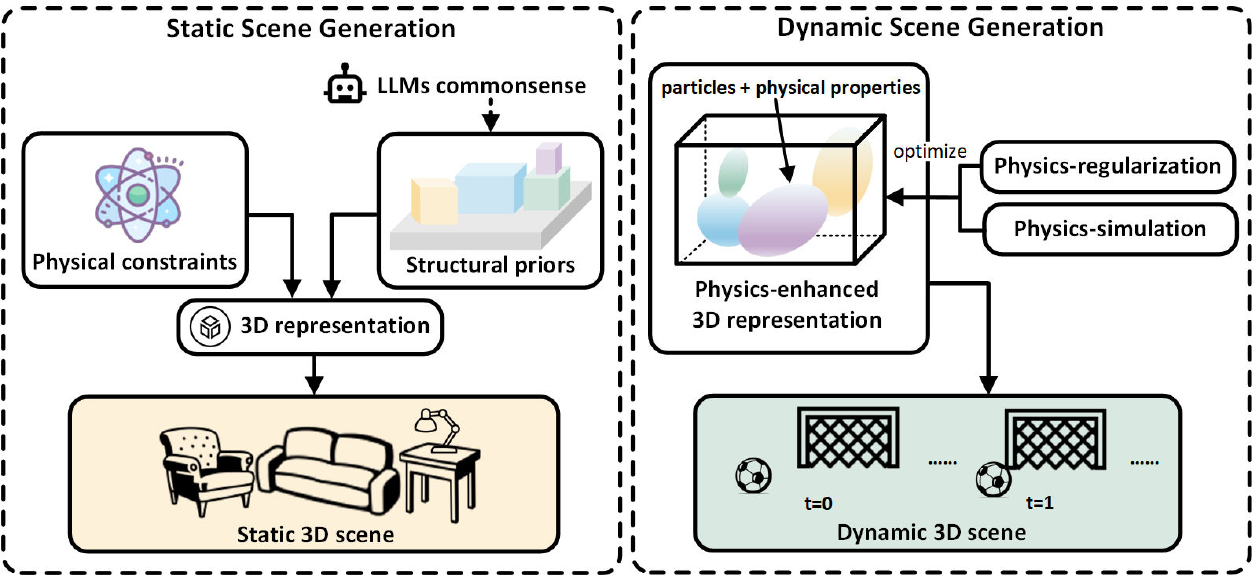}
\caption{\textbf{Illustration of the role of world knowledge in generating static and dynamic physical scenes.} Left: Static scene generation leverages structural priors from LLM commonsense and explicit physical constraints to produce plausible layouts. Right: Dynamic scene generation builds on physics-enhanced 3D representations, where particles with physical properties are optimized via simulation or regularization to ensure temporal consistency and physical realism.}
\label{fig:generation}
\end{figure*}

In summary, as a core capability of 3D world modeling, 3D physical scene generation advances world models from visually realistic renderers to physically grounded, cognitive systems. Recent developments reveal a clear trajectory toward unifying 3D representations with physically grounded world knowledge, as highlighted by Fig.~\ref{fig:generation}. By incorporating spatial priors, physical constraints, and material-aware dynamics into neural generation frameworks, these methods enable physically plausible synthesis of both static and dynamic environments. 

\section{3D Spatial Reasoning}
\label{sec:reasoning}

3D spatial reasoning is essential for enabling world models to interpret and analyze spatial structure, semantic context, and dynamic changes of 3D environments, thereby supporting downstream tasks such as planning and decision-making. This capability requires high-level understanding grounded in volumetric representations and contextualized by world knowledge. 

In this section, we categorize 3D spatial reasoning into two complementary dimensions: static spatial understanding and dynamic spatial analysis. The former focuses on spatial relationships and semantics within static scenes, while the latter emphasizes forecasting environmental evolution and object trajectories over time. Both research lines reflect a growing trend toward integrating 3D perception with spatial commonsense and semantic priors via LLMs and VLMs, enabling semantic inference and high-level spatial comprehension.

\subsection{Static Spatial Understanding}

Recent works in static spatial understanding have explored integrating pretrained LLMs and VLMs with 3D representations including point clouds, NeRF, and 3DGS to enhance spatial inference within 3D environments. Table~\ref{tab:static_reasoning} summarizes recent representative methods and presents their use of large pretrained models and supported reasoning tasks.

\setlength{\tabcolsep}{4.5pt}           
\renewcommand{\arraystretch}{1.1}     

\begin{table*}[!t]
  \centering
  \caption{Overview of representative works for static scene reasoning.}
  \label{tab:static_reasoning}
  \scriptsize
  \begin{tabular}{
    @{}
    >{\centering\arraybackslash}c 
    >{\centering\arraybackslash}c 
    >{\centering\arraybackslash}c
    >{\centering\arraybackslash}c
    *{6}{>{\centering\arraybackslash}c}@{}
  }
    \toprule
    \multirow{2}{*}{\textbf{Method}} & \multirow{2}{*}{\textbf{3D Rep.}} & \multirow{2}{*}{\textbf{LLMs}} & \multirow{2}{*}{\textbf{VLMs}}
      & \multicolumn{6}{c}{\textbf{Reasoning tasks}} \\
    \cmidrule(l){5-10}
     & & & & \textbf{Cap.} & \textbf{QA} & \textbf{Obj. Grd.} & \textbf{Obj. Det.} & \textbf{Seg.} & \textbf{Gen.} \\
    \midrule
    PointLLM~\cite{pointllm}            & Point cloud & \checkmark &            & \checkmark &           &           &           &           &           \\
    \midrule
    Chat‑3D~\cite{chat-3d}             & Point cloud & \checkmark &            & \checkmark & \checkmark &           &           &           &           \\
    \midrule
    3D‑LLM~\cite{3dllm}                & Point cloud & \checkmark &            & \checkmark & \checkmark & \checkmark &           &           &           \\
    \midrule
    Chat‑Scene~\cite{chat-scene}      & Point cloud & \checkmark &            & \checkmark & \checkmark & \checkmark &           &           &           \\
    \midrule
    LL3DA~\cite{ll3da}                & Point cloud & \checkmark &            & \checkmark & \checkmark &           &           &           &           \\
    \midrule
    Reason3D~\cite{reason3d}          & Point cloud & \checkmark &            & \checkmark & \checkmark & \checkmark & \checkmark & \checkmark &           \\
    \midrule
    SpatialLM~\cite{spatiallm}        & Point cloud & \checkmark &            &            &           & \checkmark & \checkmark &           &           \\
    \midrule
    GPT4Point~\cite{gpt4point}        & Point cloud & \checkmark & \checkmark  & \checkmark & \checkmark &           &           &           & \checkmark \\
    \midrule
    Grounded 3D‑LLM~\cite{grounded3dllm} & Point cloud & \checkmark &            & \checkmark & \checkmark & \checkmark & \checkmark & \checkmark &           \\
    \midrule
    LERF~\cite{lerf}                   & NeRF        &            & \checkmark &            &           & \checkmark &           &           &           \\
    \midrule
    LLaNA~\cite{llana}                 & NeRF        & \checkmark &            & \checkmark & \checkmark &           &           &           &           \\
    \midrule
    LangSplat~\cite{langsplat}        & 3DGS        &            & \checkmark &            &           & \checkmark &           &           &           \\
    \midrule
    ChatSplat~\cite{chatsplat}        & 3DGS        & \checkmark &            & \checkmark & \checkmark &           &           &           &           \\
    \midrule
    ReasonGrounder~\cite{reasongrounder} & 3DGS        &            & \checkmark &            &           & \checkmark &           &           &           \\
    \midrule
    Query3D~\cite{query3d}            & 3DGS        & \checkmark &            &            &           &           &           & \checkmark &           \\
    \midrule
    SplatTalk~\cite{splattalk}        & 3DGS        & \checkmark & \checkmark  &            & \checkmark &           &           &           &           \\
    \bottomrule
  \end{tabular}
  \\ {\raggedright\footnotesize * Abbreviations: 3D Rep.=3D Representation, Cap.=Captioning, QA=Question Answering, Obj. Grd.=Object Grounding, Obj. Det.=Object Detection, Seg.=Segmentation, Gen.=Generation.\par}
\end{table*}

\subsubsection{Point-language alignment for point cloud understanding} 
\begin{figure*}[!t]
\centering
\includegraphics[width=0.6\linewidth]{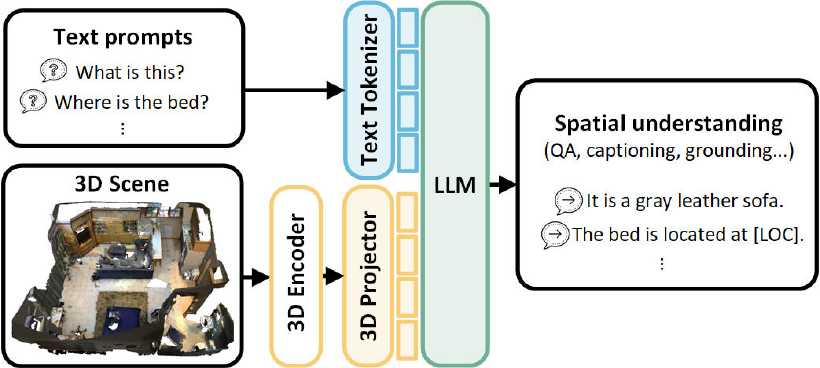}
\caption{\textbf{Illustration of the point–language alignment paradigm for 3D scene understanding.} Recent methods leverage pretrained LLMs to align point cloud features with text tokens, enabling a unified representation space that incorporates both geometric and semantic information. Point–language models like Reason3D~\cite{reason3d}, Grounded 3D-LLM~\cite{grounded3dllm} extend this approach from single-object captioning to complex scene-level reasoning tasks. }
\label{fig:reasoning}
\end{figure*}

A large number of recent methods transform unstructured point cloud data into tokenized spatial features, which are then aligned with pretrained LLMs to achieve holistic understanding, as illustrated by Fig.~\ref{fig:reasoning}. Notably, PointLLM\cite{pointllm} encodes geometry and color information of input point clouds into latent feature vectors, projects them into the embedding space of a pretrained LLM, and fuses these spatial features with text tokens. By leveraging the commonsense priors embedded in the LLM, PointLLM is capable of producing plausible and informative textual descriptions grounded in 3D geometry. 

This paradigm has been extended beyond single-object captioning to full-scene understanding\cite{sqa3d}. Chat-3D\cite{chat-3d} introduces 3D object segmentation and explicit spatial relation modeling. It first extracts object-level embeddings and then applies a relation module to capture inter-object spatial relationships. During training, it aligns both object-centric and scene-level features with a pretrained LLM in a staged manner, facilitating a progressively transition from understanding single-object attributes to modeling complex multi-object spatial relations. 3D-LLM\cite{3dllm} and Chat-Scene\cite{chat-scene} extract multi-view semantic features and visual cues from rendered images of the input 3D scene, which are combined with 3D point embeddings to construct semantically enriched spatial representations. LL3DA\cite{ll3da} aggregates multimodal prompts, including user clicks, 3D box annotations, and textual instructions, along with global scene embeddings to form unified query tokens. By aligning user intentions with rich 3D representations, LL3DA demonstrates the potential of visual-instruction tuning for human-in-the-loop 3D reasoning and decision-making. 

Builds upon the point–language alignment strategy, some methods broaden the scope of scene point cloud understanding from question answering (QA) and captioning to encompass tasks such as fine-grained spatial analysis. Reason3D\cite{reason3d} guides the LLM backbone to generate structured outputs with task-specific control tokens, enabling seamless integration with dedicated downstream modules for hierarchical object search and segmentation. SpatialLM\cite{spatiallm} extracts dense spatial features from raw point cloud data, acquired from monocular videos, RGB-D sensors, or LiDAR, and interprets them using a pretrained LLM to produce fine-grained, structured 3D scene representations. These representations include semantic categories, oriented object bounding boxes, and architectural elements such as walls, doors, and windows, thereby supporting 3D structural layout generation and agent action planning. GPT4Point\cite{gpt4point} and Grounded 3D-LLM\cite{grounded3dllm} propose two-stage frameworks to support various 3D reasoning tasks. They first performs point–text feature alignment through pretraining tasks designed to bridge geometric and linguistic modalities. Subsequently, they employ an LLM for multi-task instruction fine-tuning based on the aligned point cloud features. 

\subsubsection{Language-Guided Reasoning over NeRF and 3DGS} In addition to point-cloud-based approaches, researchers also investigate distilling knowledge from pretrained LLMs or VLMs into NeRF\cite{lerf,connectnerf,llana} and 3DGS\cite{chatsplat, reasongrounder, query3d, splattalk}, so that these 3D formats can be queried and analyzed via textual prompts. 

To support object grounding in NeRF scenes, LeRF learns a 3D language field by associating each spatial location with a CLIP embedding. It performs multi-scale volume rendering of these CLIP features along camera rays, aligning them with multi-view, multi-scale image-level CLIP embeddings. Additionally, a DINO-based regularization is applied to improve object boundary sharpness and ensure view consistency. LangSplat\cite{langsplat} and LE3DGS\cite{le3dgs}extend this approach by leveraging 3DGS for faster rendering and improved semantic precision. They further incorporate hierarchical semantics via SAM-generated masks to refine object boundary delineation. Query3D\cite{query3d} augments LE3DGS by introducing LLM-generated language queries to enrich the language embedding space, thereby improving open-vocabulary querying performance. ReasonGronder\cite{reasongrounder} further advances this line of work by mapping both language-aligned and instance-level features into the 3DGS space. It supervises language-aligned features with multi-view CLIP embeddings and distills instance-aware features from SAM masks using a contrastive loss. his hierarchical supervision boosts grounding accuracy in occluded regions and enhances generalization to implicit, open-vocabulary queries.

Similar to the point-language models, some approaches propose to tokenize NeRF or 3DGS representations, and aligns them with pretrained LLMs for question answering. LLaNa\cite{llana} directly ingests the parameters of a trained NeRF MLP. Specifically, it introduces a meta-encoder (nf2vec) that compresses the raw network weights into a compact latent vector, which is injected as prefix tokens into a frozen LLM. This alignment allows the model to interpret and reason about the encoded 3D scene using textual prompts. ChatSplat\cite{chatsplat} encodes view-, object-, and scene-level features rendered from 3DGS and converts them into input tokens of a LLM, to achieve multi-level chatting over the 3D scene. SplatTalk\cite{splattalk} introduces a framework for 3D visual question answering over 3DGS representation. It begins by constructing a 3DGS scene from a set of posed RGB images and enriches each Gaussian with semantic features extracted from VLMs. These enriched Gaussians are then converted into 3D tokens, which are fed into a pretrained LLM for open-ended spatial reasoning. 

\subsection{Dynamic Spatial Analysis}

Beyond static spatial understanding, 3D spatial reasoning also involves dynamic spatial analysis. This entails modeling the temporal evolution of the 3D scene and forecasting its future states. 

Classic dynamic spatial analysis methods convert multi-source sensor data, including cameras, LiDAR, and radar, into Bird's-Eye View (BEV) representations to capture spatiotemporal information for detecting and predicting dynamic objects\cite{bev, tsbev}. BEV representations inherently compress vertical geometry and occlude 3D structural details, limiting their capacity to characterize vertical structures and depict fine-grained internal geometric shapes of objects\cite{surveyocc}. Modern approaches represent dynamic scenes as a series of 3D occupancy grids, which preserve full volumetric geometry of the scene over time\cite{occ3d, occnet, surroundocc}. Given a historical multi-view image sequence or LiDAR sequence, the goal is to learn a 3D semantic occupancy field for each time step.

Recent methods further enhance occupancy-based dynamic scene analysis by introducing semantic priors from VLMs, i.e., pretrained open-vocabulary segmentation models\cite{grounding-dino, sam, groundedsam}, and employing neural rendering for self-supervision, thereby reducing reliance on 3D annotations. OccNeRF\cite{occnerf} parameterizes the occupancy field and employs neural rendering to generate multi-view depth maps, supervised by multi-frame photometric consistency. It further leverages Grounding DINO\cite{grounding-dino} and SAM\cite{sam} to guide semantic label prediction. OccGS\cite{occgs} extracts semantic information using these VLMs, and align it with LiDAR points to construct semantic- and geometric-aware Gaussian representations of dynamic scenes. It enables zero-shot, semantically rich modeling of dynamic environments. 

Unlike the above-mentioned methods that predict per-frame occupancy independently, Let Occ Flow\cite{letoccflow} introduces occupancy flow fields to capture temporal voxel-level motion in the horizontal plane, enabling explicit modeling of object dynamics. The method leverages semantic and optical flow cues extracted from the pretrained Grounded SAM\cite{groundedsam} model, enabling effective dynamic decomposition and providing supervision for learning motion patterns.

In summary, 3D spatial reasoning equips world models with the capability to comprehend both the static and dynamic aspects of 3D world. Recent efforts enrich 3D geometric representations with world knowledge encoded in pretrained LLMs or VLMs, enabling reasoning grounded in both spatial structure and semantic context.

\section{3D Spatial Interaction}
\label{sec:interaction}

3D spatial interaction equips world models with the ability to solve interactions with agents or humans, and manipulate 3D environments in a physically grounded and goal-directed manner.

In this section, we categorize active 3D interaction methods according to the actor’s perspective and role within the interaction process: egocentric embodied interaction and exocentric scene manipulation. The former refers to scenarios where an embodied agent interacts with the environments, navigating, manipulating objects, or executing tasks based on physical and semantic understanding. The latter involves user-driven interaction with the scene from an external or top-down perspective, where users edit, rearrange, or transform elements within the environment to achieve desired outcomes.

\subsection{Egocentric Embodied Interaction}

\begin{figure*}[!t]
\centering
\includegraphics[width=0.7\linewidth]{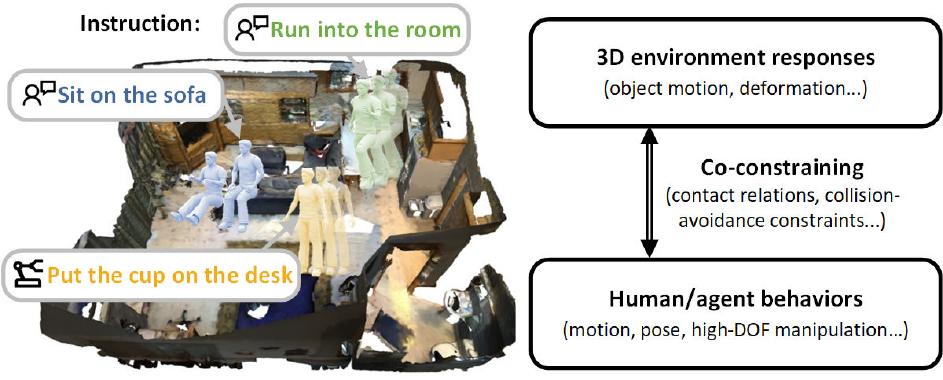}
\caption{\textbf{Illustration of egocentric embodied interaction.} A key challenge lies in capturing the co-constraining relationship between human/agent's behaviors and the environment’s responses, where actions like sitting, placing objects, or navigating space should dynamically respect contact relations and collision constraints.}
\label{fig:interaction}
\end{figure*}

This category focuses on modeling egocentric interactions where a human or an agent engages within the 3D environment, by predicting physically and semantically plausible actions and corresponding environment responses, as illustrated in Fig.~\ref{fig:interaction}.

\subsubsection{Human-scene interaction} methods synthesize physically plausible human motions by aligning language instructions, human motion, and 3D scene context. HUMANISE\cite{humanise} introduces a large-scale synthetic dataset of language-grounded human–scene interactions and proposes a generative model that produces SMPL\cite{smpl}-based full body motions conditioned on textual prompts and scene geometry. The model learns the joint embedding of the 3D scene and the language description as the condition to generate human motion. It employs auxiliary losses to enforce accurate object localization and action-specific motion generation, enhancing both physical plausibility and semantic coherence. TRUMANS\cite{trumans} extends this idea to real-world data by contributing a motion capture dataset paired with scanned indoor scenes. Its diffusion-based model synthesizes contact-aware motion from 3D layouts using local occupancy perception. At each step, a local scene perceiver queries global occupancy to maintain 3D-aware collision avoidance, while action progression embeddings condition the diffusion process, achieving real-time and generalizable egocentric behavior.

Recent works have increasingly investigated bidirectional agent-environment interactions, acknowledging that agent behaviors and environmental configurations are interdependent and co-constraining, as illustrated by Fig~\ref{fig:interaction}. CHOIS\cite{chois} presents a conditional diffusion framework that synchronizes human–object interactions under sparse waypoints and natural language instruction, and designs guidance terms to enforce contact constraints between actors and objects. For example, it can follow the textual instructions, such as “move the desk lamp next to the sofa”, to generate corresponding synchronized human-object motions while accurately handling hand-object contact and object-floor collision avoidance. ZeroHSI\cite{zerohsi} further enhances the representation of human-scene interaction using heterogeneous 3D Gaussians to handle object transformation and occlusion. It proposes a combination of the parameterized Gaussian human, transformed Gaussian object and static Gaussian scene, and joint optimizes egocentric camera pose, body motion, and object motion. It refines body motion using a pretrained pose prior derived from human motion-capture data, and employs hand-object contact and human pose smoothness losses to enforce physically consistency of interaction sequences.

\subsubsection{Agent-scene interaction} focuses on predicting precise, high-degree-of-freedom (high-DOF) manipulation actions in 3D environments. Recent methods reconstruct voxel- or point cloud-based workspace representations, integrate pretrained language models to interpret instructions and embed spatial commonsense, then combine spatial and linguistic features to predict actions such as translation, rotation, and gripper state. Perceiver-Actor\cite{peract} reconstructs a voxel grid from RGB-D sensor inputs and encodes language instructions using a pretrained language model. It then extracts RGB-D voxel patches and fuses semantic and geometric information through cross-modal attention. Finally, it predicts a sequence of continuous actions using a voxel decoder, formulating action selection as a per-voxel classification task. PolarNet\cite{polarnet} proposes a point cloud-based approach that combines a point cloud encoder with a CLIP text encoder by integrating their intermediate features through a multi-layer transformer. A point cloud decoder is then employed to predict scene and agent transitions, including per-point position offsets, as well as the gripper's rotation and open state. 3DAPNet\cite{3dapnet} adopts a point cloud-based framework that encodes language instructions with a pretrained transformer and jointly models object affordance region and affordance-conditioned action poses, enabling fine-grained and instruction-aligned manipulation.

\subsection{Exocentric Scene Manipulation}

Interactive scene manipulation refers to user-initiated interaction with 3D environments from an external viewpoint, where users edit elements of a scene to achieve specific goals. These approaches build on neural 3D representations, notably NeRFs or 3DGS, as editable substrates and often leverage language prompts, pixel inputs, or other visual cues to guide interactive editing.

\setlength{\tabcolsep}{3pt}           
\renewcommand{\arraystretch}{1.1}     

\begin{table*}[!t]
  \centering
  \caption{Overview of representative works for exocentric scene manipulation}
  \label{tab:interaction}
  \scriptsize
  \begin{tabular*}{\textwidth}{@{\extracolsep{\fill}}%
    m{0.12\textwidth}
    m{0.21\textwidth}
    m{0.06\textwidth}
    m{0.10\textwidth}
    m{0.21\textwidth}
    m{0.20\textwidth}
  @{}}
    \toprule
    \textbf{Category} & \textbf{Method} & \textbf{3D Rep.} & \textbf{Prompts} & \textbf{Priors} & \textbf{Applicability} \\
    \midrule
    \multirow{8}{=}{Static scene manipulation}
      & Instruct- NeRF2NeRF~\cite{instructnerf}
        & NeRF & Text & 2D diffusion prior
        & Contextual edition, object stylization \\
    \cmidrule{2-6}
      & SIn-NeRF2NeRF~\cite{sinnerf}
        & NeRF & Text & 2D diffusion prior, segmentation\&inpainting
        & Object translation, rotation, scaling \\
    \cmidrule{2-6}
      & CLIP-NeRF~\cite{clipnerf}
        & NeRF & Text/image & Text-visual prior via CLIP
        & Object stylization \\
    \cmidrule{2-6}
      & GaussianEditor~\cite{gaussianeditor}
        & 3DGS & Text & Diffusion prior, LLM-guided segmentation
        & Region-level removal, styling \\
    \cmidrule{2-6}
      & Point'n Move~\cite{pointnmove}
        & 3DGS & 2D click & Segmentation \& inpainting
        & Object translation, rotation, removal \\
    \midrule
    \multirow{7}{=}{Dynamic scene manipulation}
      & Instruct 4D-to‑4D~\cite{instruct4d24d}
        & NeRF & Text & 2D diffusion prior
        & Scene stylization \\
    \cmidrule{2-6}
      & 4D-Editor~\cite{4deditor}
        & NeRF & User’s stroke & Semantic prior via DINO
        & Object translation, removal, recoloring \\
    \cmidrule{2-6}
      & Neuphysics~\cite{neuphysics}
        & SDF & Physics parameters & Differentiable physics engine
        & Object removal, duplication, recoloring \\
    \cmidrule{2-6}
      & SC‑GS~\cite{scgs}
        & 3DGS & Control points & ARAP rigidity constraint
        & Object motion edition \\
    \bottomrule
  \end{tabular*}
  {\raggedright\footnotesize * Abbreviations: 3D Rep. = 3D Representation. \par}
\end{table*}

\subsubsection{Static scene manipulation} has increasingly leveraged text and image prompts to enable intuitive and fine-grained manipulations. A series of NeRF-based approaches integrate pretrained 2D diffusion models or vision-language priors, achieving semantic coherent edits. Instruct-NeRF2NeRF\cite{instructnerf} proposes a text-driven NeRF editing pipeline that iteratively edits the training images of a NeRF using a pretrained 2D image-conditioned diffusion model, and re-optimizes the NeRF for multi-view consistency. Building on this, SIn-NeRF2NeRF\cite{sinnerf} enhances object-specific manipulation by disentangling object and background regions using multi-view segmentation and background inpainting, supporting object-level transformations like translation and resizing. CLIP-NeRF\cite{clipnerf} distills open-vocabulary visual concepts from CLIP\cite{clip} into a disentangled conditional NeRF architecture. Given a text prompt or reference image, CLIP encoders extract shape and appearance cues, which guide the NeRF to perform targeted edits with separate control over geometry and texture. A CLIP-based similarity loss is applied to ensure that the edited NeRF maintains multi-view semantic consistency with the input prompts. 

Recent methods extend manipulation to 3DGS-based representations for improved efficiency and editability. GaussianEditor\cite{gaussianeditor} introduces a text-driven 3DGS editing framework that supports fine-grained, region-level manipulation. It first employs an LLM to identify regions of interest (ROI) based on the scene description and user instructions. These textual ROIs guide image-space segmentation via Grounding-DINO and SAM, which is then lifted to 3D space. A pretrained text-to-image diffusion model generates edited images from the instructions, serving as supervision to optimize the corresponding 3D Gaussians, thus enabling semantically consistent and localized 3D edits. 

Beyond text prompts, interactive approaches empower users with direct control over object manipulation. Point’n Move\cite{pointnmove} presents an interactive 3DGS editing framework based on user's 2D clicks. Its two-stage self-prompting segmentation strategy first associates user clicks with the appropriate 3D object regions, then propagates the segmentation across multiple views. Upon object selection, the system supports direct manipulations such as translation, rotation, removal by inpainting exposed regions in the 3DGS representation. 

\subsubsection{Dynamic scene manipulation} poses additional challenges, such as maintaining spatiotemporal coherence and handling object motion. Instruct 4D-to-4D~\cite{instruct4d24d} treats a dynamic 4D scene as a collection of pseudo-3D videos. It adopts a pretrained 2D diffusion model augmented with anchor-aware attention to batch-edit key frames consistently. The edits are then propagated both temporally via optical-flow warping and spatially via depth-based projection to other views, achieving sharp, spatially and temporally coherent 4D scene edits. 4D-Editor\cite{4deditor} supports stroke-based editing on dynamic NeRFs. It incorporates semantic priors distilled from a pretrained DINO\cite{dino} teacher model and recursive refinement to enhance object tracking and segmentation across time, and employs multi-view temporal inpainting to ensure spatiotemporal coherence. 

Beyond appearance editing, some methods focus on physically plausible dynamic manipulation. NeuPhysics\cite{neuphysics} proposes a physics-based framework for dynamic scene manipulation from monocular RGB videos. It represents dynamic scenes using a time-varying signed distance field (SDF) and jointly estimates geometry, appearance, motion, and physics parameters through a differentiable physics engine. By modifying these parameters, it enables dynamic object editing, such as removal, duplication, or recoloring, while maintaining physical plausibility and temporal coherence. For motion-level control, SC-GS\cite{scgs} enables motion editing of dynamic objects by introducing a set of sparse control points, which guide the deformation of Gaussians over time via a learned deformation MLP. To preserve physical plausibility and spatial continuity, it incorporates an As-Rigid-As-Possible (ARAP) loss that enforces local rigidity during deformation.

In summary, 3D spatial interaction enables world models to engage with and reshape 3D environments. Egocentric embodied interaction focuses on modeling plausible human/agent behaviors and environment responses, while exocentric scene manipulation emphasizes user-guided scene editing through language, visual, or spatial inputs. Together, these approaches highlight the growing importance of interaction-aware 3D representations that integrate physical constraints, semantic priors, and intuitive control interfaces, laying the groundwork for more intelligent and responsive 3D world models.

\section{Application}
\label{sec:application}

World models are powering a wide range of real-world applications, including embodied AI, autonomous driving, digital twins, and gaming/VR. Each of these domains presents unique requirements and constraints. In this section, we investigate the recent advancements of specialized models and platforms that combine the core cognitive capabilities of 3D world modeling to tackle domain-specific challenges and support complex tasks.

\subsection{Embodied AI}
\label{sec:robotics}

\begin{figure*}[!h]
\centering
\includegraphics[width=0.9\linewidth]{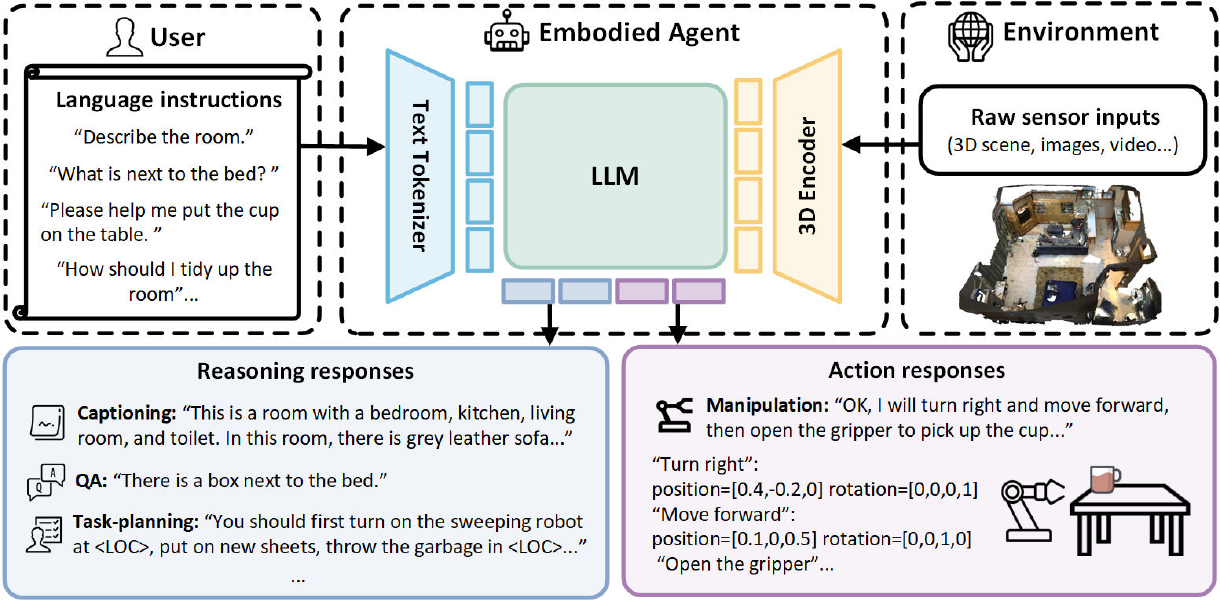}
\caption{\textbf{Representative paradigm of end‑to‑end generative world models for embodied AI.} The agent receives raw sensor inputs from the environment and language instructions from the user, and performs alignment using LLMs to generate reasoning and action responses for tasks such as captioning, QA, and task-planning. End-to-end models such as 3D-VLA\cite{3dvla} and LEO\cite{leo} leverage this framework to enable intelligent agents to interpret, reason, and act in complex real-world scenes.}
\label{fig:embodied}
\end{figure*}

In embodied AI systems, agents must form a deep 3D understanding of their surroundings, by reasoning about spatial relationships, predicting plausible actions, to succeed in complex tasks under real‐world scenarios.

Recent end‑to‑end generative world models for embodied AI unify perception, reasoning, and action within a shared 3D spatial and semantic representation, enabling agents to understand and interact with complex environments, as illustrated in Fig~\ref{fig:embodied}. 3D-VLA\cite{3dvla} introduces a generative vision-language-action model. It builds on a pretrained 3D-centric LLM backbone augmented with interaction tokens, which ingests both textual instructions and 3D observations in a shared embedding space to understanding 3D states and user' intentions. Then, a series embodied diffusion models are trained and their latent features are aligned with the 3D-LLM’s representation, thus enabling the model to reconstruct current scenes and imagine goal states in form of both RGB-D and point clouds. During inference, given a natural language command, 3D-VLA alternates between invoking the diffusion decoders to predict future observations and using the 3D-LLM to decode corresponding action tokens, effectively closing the loop from perception through prediction to planning in a single forward pass. LEO \cite{leo} is a 3D embodied multimodal world models for perception, grounding, reasoning, planning, and action in 3D environments. It first achieves 3D vision–language alignment by using a 2D image encoder for egocentric views and a 3D point-cloud encoder for global views. The resulting visual tokens, interleaved with text tokens, form a unified multimodal sequence that is fed into a decoder-only LLM. Then, extending the vocabulary to include both text tokens and discrete action tokens reframes tasks as sequence prediction, enabling end-to-end instruction following across diverse tasks. AETHER\cite{aether} realizes a unified world modeling framework by co-optimizing 4D dynamic reconstruction, action-conditioned future prediction, and goal-conditioned action planning. Its reconstruction head recovers temporally consistent 3D scene geometry and motion fields from input video sequences. Then, an action-conditioned predictor is trained to forecast future frames given past observations and candidate camera or actor trajectories, leveraging task-interleaved feature learning to share representations between reconstruction and prediction tasks. Finally, a goal-conditioned planner head maps predicted scene changes to camera paths and motion commands. Crucially, losses for reconstruction, prediction, and planning are back-propagated through a shared backbone, enabling jointly optimization and synergistic knowledge transfer. GenEx\cite{genex} constructs an explorable 360° panoramic environment from a single RGB image via diffusion-based world generator trained on large-scale, physics-grounded Unreal Engine data. GenEx allows the agent to explore the synthesized panoramic environment and uses the same diffusion model to forecast unseen regions. These imagined observations help the GPT-based planner to update agent's internal understanding of its surroundings and generate navigation or manipulation commands. In this way, GenEx performs a perception–prediction–planning loop within a unified framework, without relying on an external motion planner.

Overall, this emerging paradigm paves the way toward more capable and generalizable agents. Future efforts might focus on continual adaptation of internal 3D representations during real-world operation, and seamless integration of learned world models with reactive control loops to enable truly generalist, high-DOF robotic autonomy.

\subsection{Autonomous Driving}
\label{sec:driving}

World models for autonomous driving demand real-time performance, long-horizon consistency, and multi-sensor fusion to manage complex traffic environments\cite{driving_survey}. Recent advances leverage 3D representations, especially 3D occupancy grids and point clouds, enriched with road priors and spatiotemporal modeling, enabling prediction, planning, and question answering in dynamic driving scenarios.

OccWorld\cite{occworld} proposes 3D semantic occupancy grids as its core scene representation, capturing both static infrastructure and dynamic elements. It employs a self-supervised VQVAE “scene tokenizer” to convert dense occupancy grids into discrete tokens encoding structural and semantic information. A GPT-style spatiotemporal transformer then performs spatial mixing for multi-scale context understanding and causal temporal attention for future prediction, jointly forecasting scene evolution and ego-vehicle trajectories. OccLLaMA\cite{occllama} extends OccWorld by integrating vision, language, and action within a shared token framework. It combines VQVAE-generated scene tokens with language and action vocabularies, processed jointly by LLaMA, enabling multiple tasks such as occupancy forecasting, motion planning, and visual question answering. 

OccSora\cite{occsora} and DynamicCity\cite{dynamiccity} further explore 4D generative modeling of long-term driving scene evolution. OccSora leverages a diffusion-based Transformer on compressed temporal tokens for long-horizon generation. DynamicCity\cite{dynamiccity} introduces HexPlane VAE and DiT diffusion to synthesize high-fidelity future LiDAR or occupancy scenes conditioned on ego trajectories or layout constraints. In parallel, DFIT-OccWorld\cite{dfitoccworld} decomposes scene generation into static occupancy and dynamic flow, integrating camera-derived features via differentiable rendering to strengthen multi-sensor fusion and spatiotemporal coherence, enabling more robust and accurate voxel-level predictions. NeMo\cite{nemo} employs a two-stage, end-to-end sensorimotor framework that learns occupancy and motion flow via neural rendering and fuses current and predicted volumetric features for action planning.

On the point-cloud side, Copilot4D\cite{copilot4d} formulates an unsupervised world model on raw LiDAR point clouds. It tokenizes raw LiDAR into discrete tokens using VQVAE and models future states with a discrete diffusion Transformer yielding high-fidelity and physically consistent point-cloud forecasts for autonomous planning. ViDAR\cite{vidar} introduces a self-supervised pretraining framework termed visual point cloud forecasting. Rather than modeling point clouds directly, ViDAR predicts future LiDAR scans from historical camera images. It projects dense visual features into 3D space and autoregressively generates future point clouds, enabling the joint learning of semantic, geometric, and temporal representations in a self-supervised manner.

Despite employing different 3D backbones, these world models all unify static and dynamic scene elements into a single latent space, embed road-environment priors to ground predictions in physically plausible structures, and leverage spatiotemporal modeling to ensure long-horizon consistency and capture scene evolution. Future work will focus on integrating richer semantic information, tighter multi-sensor fusion, and more efficient 3D representations to further improve robustness and adaptability in real-world driving.

\subsection{Digital Twins}
\label{sec:twins}

Digital twins require precise digital replicas of real-world entities, encompassing both their static geometry and dynamic behaviors. Cutting-edge world models in digital twin city harness efficient 3D representations and embed urban-scene priors, such as structural layouts, road networks, and architectural semantics, to enable scalable, controllable generation of expansive urban environments.

CityDreamer\cite{citydreamer} establishes a compositional framework for unbounded city generation. Operating on a bird’s‑eye structural layout, it decomposes urban scenes into background neural fields (roads, greenery) and building instance fields. It uses generative hash grids and periodic embeddings to ensure consistency in geometry and texture while supporting diverse building appearances. CityGaussian\cite{citygaussian} builds on this by focusing on efficient, real-time rendering of large cities via 3DGS. It trains a global Gaussian prior to capture overall geometry and appearance and partitions the scene into adaptive blocks. At runtime, Gaussian splats are selected and merged based on view distance, ensuring smooth visual continuity and sustaining interactive frame rates (40–65 FPS) on unbounded cityscapes. CityDreamer4D\cite{citydreamer4d} enhances static city models by explicitly modeling dynamic entities like vehicles. It decouples traffic scenario generation from static layout creation and utilizes three specialized NeRF-based generators to separately model the static city background, buildings, and dynamic vehicles at each time step. This design effectively captures the diversity of different elements within the city, while also supporting local editing of individual building and vehicle instances, enabling fine-grained control and realistic scene composition.

Procedural controllable generation frameworks automate and customize city creation. SceneX\cite{scenex} and CityCraft\cite{citycraft}, harness LLMs to translate high‑level directives into executable scene‑creation workflows including asset placements and layout generation. Specifically, CityCraft generates 2D layouts via a diffusion Transformer and uses LLM guidance and asset retrieval tools for diverse, rich urban scenes. CityX\cite{cityx} further integrates multimodal inputs, including OpenStreetMap (OSM), semantic maps, satellite imagery, and modular procedural plugins for real‑time, iterative city editing. UrbanWorld\cite{urbanworld} presents an end‑to‑end generative urban world model that automatically constructs customizable 3D urban environments under flexible control conditions. It first generates coarse 3D layouts from OSM, utilizes an urban-specialized multimodal LLM for asset placement, and applies progressive 3D diffusion to refine geometry and textures. A final LLM‑guided polishing step aligns global scene consistency. The result is an interactive 3D urban world navigable by embodied agents.

Collectively, these digital twin world models highlight advancements toward data-driven, controllable, and scalable generation of realistic urban environments. By combining efficient 3D representations with domain priors such as urban layouts and architectural semantics, they enable high-fidelity modeling of both static structures and dynamic agents. Future work will likely explore deeper integration of multimodal inputs, enhanced interactivity, and stronger physical and behavioral consistency to support a broader range of simulation, planning, and decision-making tasks in digital twin cities.

\subsection{Gaming/VR}
\label{sec:gaming}

In gaming and VR, crafting a physically plausible, immersive, and real‑time interactive environments is essential for a compelling user experience. Traditional graphics pipelines rely on complex, costly manual modeling to develop high-quality interactive virtual worlds. Recently, 3D world models grounded in physical knowledge have transformed content generation, enabling automatic 3D scene synthesis and realistic physics-based interactions.

Video2Game\cite{video2game} presents an end-to-end pipeline that converts real-world video into interactive, photorealistic game environments. It first trains a large-scale NeRF capturing scene geometry and appearance, then distills it into a UV-mapped neural-textured mesh optimized for real-time rendering. Finally, a physics module integrates rigid-body dynamics and collision constraints, allowing virtual characters to navigate and interact realistically, such as kicking a football governed by real-world physics. VR‑GS\cite{vr-gs} is an interactive VR system that employs 3DGS alongside a sparse volumetric structure VDB to construct interactive object‑level 3D scenes. By integrating a two‑level embedding framework with XPBD\cite{xpbd}-based deformable body simulations, it enables real-time, physics-driven interactions and dynamic shadows, supporting intuitive VR gameplay such as petting virtual animals or grasping objects with natural responses. LIVE-GS\cite{livegs} constructs realistic, interactive VR environments by integrating spatial and physical commonsense through pretrained LLMs and physics simulation. It leverages GPT-4o to analyze physical properties of objects represented as 3D Gaussians, guiding the resulting dynamics to align with real-world behavior. Its PBD-based physics simulation framework supports a wide range of physical forms, including rigid, soft, and granular bodies, enabling rich, semantically grounded interactions without the need for extensive manual design.

These approaches unify 3D representations with real-world physics, enabling synthesis of interactive VR environments, realistic object behaviors, and natural user interactions. Future advances may include context-aware physics where gameplay intent dynamically shapes simulations and lightweight volumetric streaming for expansive worlds.

\section{Challenges and Breakthrough Paths}
\label{sec:challenge}

As world models progress toward achieving comprehensive 3D cognition, they encounter a series of persistent challenges that span data, modeling, and deployment. Addressing these obstacles is critical for advancing toward more robust, intelligent, and embodied world models.

\subsection{Data}
World models require jointly reasoning over text, images, video, 3D captures and other sensor streams. However, several challenges impede effective multimodal fusion:

\begin {itemize}
\item \textbf{Modality heterogeneity and semantic misalignment:} Disparities in data formats, sampling rates, and feature distributions across various modalities complicate alignment and synchronization processes. Additionally, semantic gaps between modalities hinder seamless fusion and joint reasoning, often introducing cross-modal noise that degrades the quality of the fused representation. 
\item \textbf{Data quality and robustness issues in 3D captures:} Real-world 3D data frequently suffer from sensor-induced noise, occlusions, and missing values, adversely affecting model robustness. These quality issues pose significant challenges in accurately capturing and interpreting complex environments, especially in safety-critical applications like autonomous driving.
\item \textbf{Dataset limitations and annotation constraints:} Existing datasets often underrepresent rare scenarios, such as extreme weather conditions or unusual traffic incidents, limiting model generalizability. Furthermore, the high cost of dense 3D annotation, particularly for volumetric or point cloud data, restricts the scale and diversity of available training datasets. The lack of synchronized trajectory, control, and state-interaction annotations further impedes progress toward truly interactive and temporally coherent 3D cognition.
\end {itemize}

To address these challenges, it is essential to develop standardized data formats and sampling protocols to facilitate alignment across modalities. Employing advanced fusion techniques that account for semantic and distributional differences can enhance integration. Implementing noise-reduction algorithms and robust preprocessing methods can mitigate the impact of sensor noise and occlusions. Expanding datasets to include rare and critical scenarios, possibly through data augmentation or simulation, will improve model generalizability. Investing in semi-automated or AI-assisted annotation tools can reduce the costs associated with dense 3D data labeling. Finally, incorporating synchronized trajectory and state-interaction data will enable the modeling of dynamic, interactive environments, advancing the capabilities of world models.

\subsection{Modeling}
World models aim to represent and simulate complex, dynamic environments. However, several challenges impede the development of efficient and accurate models:

\begin {itemize}
\item \textbf{Scalability and memory efficiency in 3D scene representations:} Traditional voxel grids and point-cloud networks suffer from cubic (or worse) memory growth as scene scale increases, rendering them inefficient for handling city-level modeling. While sparse structures, such as octrees and Gaussian-splatting representations, can partially alleviate memory demands, achieving an optimal balance between preserving fine-scale geometric or radiance detail and constraining overall memory consumption remains an unresolved challenge.
\item \textbf{Integration of complex physical dynamics:} Embedding accurate fluid dynamics, material deformations, and multi-object interactions into neural scene representations remains challenging. High-precision, real-time simulation of complex dynamics is still computationally prohibitive, limiting the applicability of current models in scenarios requiring detailed physical interactions.
\item \textbf{Development of real-time, multimodal instruction-to-action pipelines:} Robust pipelines for real-time instruction-to-action, which integrate voice, touch, and visual cues into spatially grounded reasoning and planning, have not been fully developed. The lack of such integrated systems hampers the ability of models to perform real-time, context-aware actions based on multimodal inputs.
\end {itemize}

In light of these limitations, the integration of geometric perception architectures with hierarchical sequence models is being explored to enhance 3D representation capabilities. By incorporating physical priors into model training, researchers aim to improve the realism and dynamic stability of generated scenes. Furthermore, advancing cross-modal fusion techniques facilitates the coordination of language, vision, and action, which is essential for developing truly interactive and responsive 3D environments.

\subsection{Deployment}
Deploying world models in real-world scenarios presents several challenges that hinder real-time, scalable, and coherent operation:

\begin {itemize}
\item \textbf{High model complexity and inference latency:} Large-scale or physics-aware representations often entail substantial computational demands, leading to inference delays that preclude fine-grained, real-time interaction in dynamic 3D environments. This latency undermines the responsiveness required for applications such as autonomous navigation and interactive simulations.
\item \textbf{Scalability and coherence in multi-agent systems:} Deploying multiple embodied agents within a shared 3D space introduces challenges related to scalability, real-time responsiveness, and the maintenance of global coherence. Ensuring consistent and synchronized behavior among agents is essential for robust real-world systems.
\end {itemize}

Several approaches have emerged to address these challenges. Implementing model compression techniques, such as pruning and quantization, alongside distributed edge computing strategies can reduce latency and computational overhead. Leveraging federated learning, multi-task learning, and transfer learning frameworks further enhances adaptability and generalization across diverse environments. Some approaches suggest that integrating causal reasoning frameworks is critical in specific domains\cite{lecuncausal}, such as embodied intelligence\cite{causality}, to enable interpretable, multi-factor decision-making and robust planning in novel scenarios.

\section{Conclusion}

This survey explores the evolution of world models, focusing on their transition from 2D visual representations to 3D cognition. By highlighting recent breakthroughs in the field of 3D representation learning, such as NeRF and 3DGS, as well as the integration of world knowledge, we elucidate the three fundamental capabilities required for achieving genuine 3D cognition: 3D physical scene generation, 3D spatial reasoning, and 3D spatial interaction. Through a detailed analysis of practical applications in embodied AI, autonomous driving, digital twins, and gaming/VR, we demonstrate how world models apply these capabilities to complex, dynamic environments. Finally, by analyzing the unresolved challenges in this rapidly evolving field, we emphasize several promising breakthrough directions. Through the synergistic innovation of data, modeling, and deployment, we envision that next-generation world models will progressively evolve into intelligent systems capable of perceiving, reasoning, and acting within richly structured 3D environments, thus bringing us closer to the ultimate goal of achieving robust, general-purpose artificial intelligence.


\bibliographystyle{ACM-Reference-Format}
\bibliography{main}


\end{document}